\documentclass[12pt,A4]{article}
\usepackage[margin=1in,footskip=0.25in]{geometry}

\usepackage{graphicx, color}  %Required
\usepackage{amsmath, amsfonts, MnSymbol, mathrsfs}
\usepackage{natbib}
\usepackage{hyperref}
\usepackage{fancyhdr}

\def\cn{\centerline} 

\begin{document}
\title{PiML Toolbox for Interpretable Machine Learning Model Development and Diagnostics}
\author{Agus Sudjianto, Aijun Zhang$^\ast$, Zebin Yang, Yu Su and Ningzhou Zeng}
\date{Dec 18, 2023}
\maketitle
	
\begin{abstract}%   
PiML (read $\pi$-ML, /`pai·`em·`el/) is an integrated and open-access Python toolbox for interpretable machine learning model development and model diagnostics. It is designed with machine learning workflows in both low-code and high-code modes, including data pipeline, model training and tuning, model interpretation and explanation, and model diagnostics and comparison. The toolbox supports a growing list of interpretable models (e.g. GAM, GAMI-Net, XGB1/XGB2) with inherent local and/or global interpretability. It also supports model-agnostic explainability tools (e.g. PFI, PDP, LIME, SHAP) and a powerful suite of model-agnostic diagnostics (e.g. weakness, reliability, robustness, resilience, fairness). Integration of PiML models and tests to existing MLOps platforms for quality assurance are enabled by flexible high-code APIs. Furthermore, PiML toolbox comes with a comprehensive user guide and hands-on examples, including the applications for model development and validation in banking. The project is available at {\sf https://github.com/SelfExplainML/PiML-Toolbox}. 

\end{abstract}

\section{Introduction}  
Supervised machine learning has being increasingly used in domains where decision making can have significant consequences. However, the lack of interpretability of many machine learning models makes it difficult to understand and trust the model-based decisions. This leads to growing interest in interpretable machine learning and model diagnostics. There emerge algorithms and packages for model-agnostic explainability, including the {\sf inspection} module (including permutation feature importance, partial dependence) in {\sf scikit-learn} \citep{pedregosa2011scikit} and various others, e.g.  \cite{kokhlikyan2020captum, klaise2021alibi, baniecki2021dalex, li2022interpretdl}.

Post-hoc explainability tools are useful for black-box models, but they are known to have general pitfalls \citep{rudin2019stop,  molnar2022general}. Inherently interpretable models are suggested for machine learning model development \citep{yang2020enhancing, yang2021gami, sudjianto2020unwrapping}. The {\sf InterpretML} package \citep{nori2019interpretml} by Microsoft Research is such a package of promoting the use of inherently interpretable models, in particular their explainable boosting machine (EBM) based on the GA2M structure \citep{lou2013accurate}. See \citep{lengerich2020purifying, yang2021gami, hu2023interpretable} for other variants of interpretable GA2M models. One may also refer to \cite{sudjianto2021designing} for discussion about how to design inherently interpretable machine learning models. 

In the meantime, model diagnostic tools become increasingly important for model validation and outcome testing. New tools and platforms are developed for model weakness detection and error analysis, e.g., \cite{chung2019slice}, PyCaret package, TensorFlow model analysis, FINRA's model validation toolkit, and Microsoft's responsible AI toolbox. They can be used for arbitrary pre-trained models, in the same way as the post-hoc explainability tools. Such type of model diagnostics or validation is sometimes referred to as black-box testing, and there is an increasing demand of diagnostic tests for quality assurance of machine learning models. 
%There is an increasing need of such type of tests for model reliability, robustness, resilience, and fairness.

It is our goal to design an integrated Python toolbox for interpretable machine learning, for both model development and model diagnostics. This is particularly needed for model risk management in banking, where it is a routine exercise to run model validation including evaluation of model conceptual soundness and outcome testing from various angles. An inherently interpretable machine learning model tends to be more conceptually sound, while it is subject to model diagnostics in terms of accuracy, fairness, weakness detection, reliability, robustness and resilience. The PiML toolbox we develop is such a unique Python tool that supports not only a growing list of interpretable models, but also an enhanced suite of diagnostic tests. It has been adopted by multiple financial institutions since its first launch on May 4, 2022.

\section{Toolbox Design}
\begin{figure}[htp!]
\cn{\includegraphics[width=0.8\textwidth]{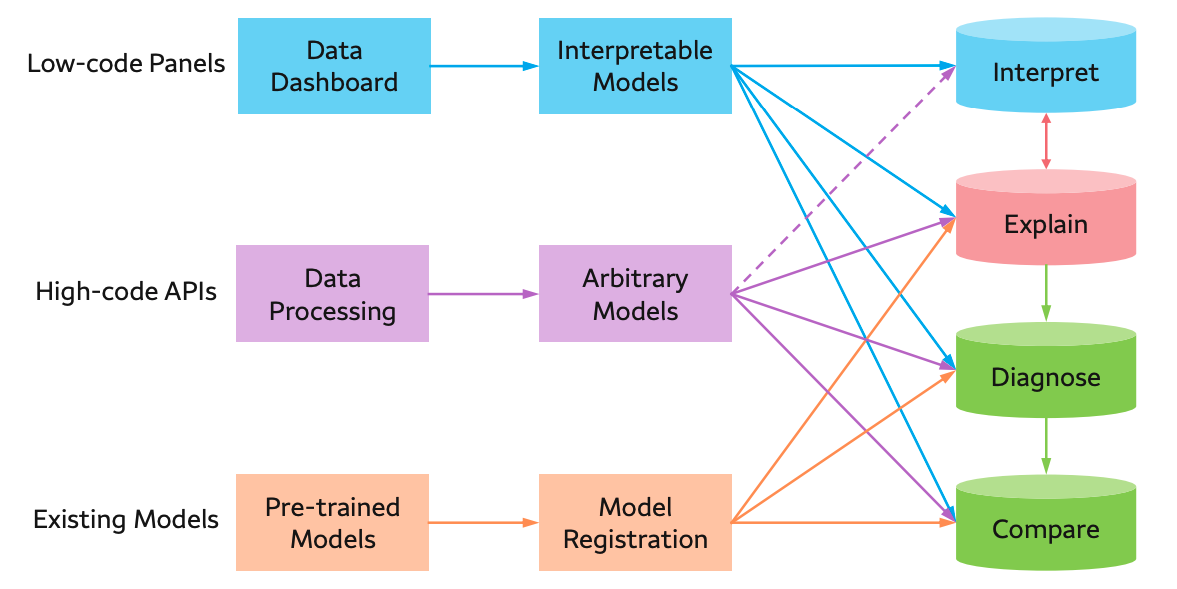}}
\caption{Design of PiML pipelines with low-code interface and high-code APIs.}\label{fig:Workflow}
\end{figure}

PiML toolbox is designed to support machine learning workflows through low-code interface and high-code APIs. It also supports registration of existing models that are pre-trained by certain other frameworks. See Figure~\ref{fig:Workflow} for the overall design of PiML pipelines. 

\begin{itemize}
    \item {\sf Low-code panels:} interactive widgets or dashboards are developed for Jupyter notebook or Jupyter lab users. A minimum level of Python coding is required. The data pipeline consists of convenient {\sf exp.data\_load(), exp.data\_summary(), exp.eda(), exp.data\_quality(), exp.feature\_select(), exp.data\_prepare()}, each calling an interactive panel with choices of parameterization and actions.   
    \item {\sf High-code APIs:} each low-code panel can be also executed through one or more Python functions with manually specified  options and parameters. Such high-code APIs are flexible to be called both in Jupyter notebook cells and by Terminal command lines. High-code APIs usually provide more options than their default use in low-code panels. End-to-end pipeline automation can be enabled with appropriate high-code settings.  
    \item {\sf Existing models:} a pre-trained model can be loaded to PiML experimentation through pipeline registration. It is mandatory to include both training and testing datasets, in order for the model to take the full advantage of PiML explanation and diagnostic capabilities. It can be an arbitrary model in supervised learning settings, including regression and binary classification.
\end{itemize}

For PiML-trained models by either low-code interface or high-code APIs, there are four follow-up actions to be executed: 
\begin{itemize}
    \item {\sf model\_interpret():} this unified API works only for inherently interpretable models (a.k.a. glass models) to be discussed in Section~3. It provides model-specific interpretation in both global and local ways. For example, a linear model is interpreted locally through model coefficients or marginal effects, while a decision tree is interpreted locally through the tree path.  
    \item {\sf model\_explain():} this unified API works for arbitrary models including black-box models and glass-box models. It provides post-hoc global explainability through permutation feature importance (PFI) and partial dependence plot (PDP) through {\sf sklearn.inspection} module, accumulated local effects \citep{apley2020visualizing}, and post-hoc local explainability through LIME \citep{ribeiro2016should} and SHAP \citep{lundberg2017unified}.  
    \item {\sf model\_diagnose():} this unified API works for arbitrary models and performs model diagnostics to be discussed in Section~4. It is designed to cover standardized general-purpose tests based on model data and predictions, i.e. model-agnostic tests. There is no need to access the model internals.  
    \item {\sf model\_compare():} this unified API is to compare two or three models at the same time, in terms of model performance and other diagnostic aspects. By inspecting the dashboard of graphical plots, one can easily rank models under comparison. 
\end{itemize}

For registered models that are not directly trained by PiML, they are treated as black-box models, even though such a model may be inherently interpretable. This is due to simplification of pipeline registration, where only the model prediction method is considered. For these models, {\sf model\_interpret()} is not valid, while the other three unified APIs are fully functional. Note that PiML since version 0.6 also supports model diagnostics based on model prediction scores, which makes a pseudo model even without access to model objects.

Regarding PiML high-code APIs, it is worthwhile to mention that these APIs are flexible enough for integration into existing MLOps platforms. After PiML installation to MLOps backend, the high-code APIs can be called not only to train interpretable models, but also to perform arbitrary model testing for quality assurance. 

\section{Interpretable Models}
PiML supports a growing list of inherently interpretable models. For simplicity, we only list the models and the references. One may refer to the PiML user guide \citep{PiMLdoc} for details of each model use and hands-on examples. The following list of interpretable models are included PiML toolbox V0.5 (latest update: May 4, 2023). 

\begin{enumerate}
\item {\sf GLM}: Linear/logistic regression with $\ell_1$ and/or $\ell_2$ regularization \citep{hastie2015statistical}

\item {\sf GAM}: Generalized additive models using B-splines \citep{serven2018pygam}

\item {\sf Tree}: Decision tree for classification and regression \citep{pedregosa2011scikit}

\item {\sf FIGS}: Fast interpretable greedy-tree sums \citep{tan2022fast}

\item {\sf XGB1}: Extreme gradient boosted trees of depth 1, using optimal binning \citep{chen2015xgboost, navas2020optimal}

\item {\sf XGB2}: Extreme gradient boosted trees of depth 2, with purified effects \citep{chen2015xgboost, lengerich2020purifying}

\item {\sf EBM}: Explainable boosting machine \citep{lou2013accurate, nori2019interpretml}

\item {\sf GAMI-Net}: Generalized additive model with structured interactions \citep{yang2021gami}

\item {\sf ReLU-DNN}: Deep ReLU networks using Aletheia unwrapper and sparsification \citep{sudjianto2020unwrapping}
\end{enumerate}

\section{Diagnostic Suite}
PiML comes with a continuously enhanced suite of diagnostic tests for arbitrary supervised machine learning models under regression and binary classification settings. Below is a list of the supported general-purpose tests with brief descriptions. One may refer to the PiML user guide \citep{PiMLdoc} and PiML tutorials \citep{PiMLMedium} for details of each diagnostic test and hands-on examples.

\begin{enumerate}
\item {\sf Accuracy}: popular metrics like MSE, MAE for regression tasks and ACC, AUC, Recall, Precision, F1-score for binary classification tasks. 

\item {\sf WeakSpot}: identification of weak regions with high magnitude of residuals by 1D and 2D slicing techniques.

\item {\sf Overfit/Underfit}: identification of overfitting/underfitting regions according to train-test performance gap, also by 1D and 2D slicing techniques. 

\item {\sf Reliability}: quantification of prediction uncertainty by split conformal prediction techniques.

\item {\sf Robustness}: evaluation of performance degradation under different sizes of covariate noise perturbation \citep{cui2023enhancing}.

\item {\sf Resilience}: evaluation of performance degradation under different out-of-distribution scenarios.

\item {\sf Fairness}: disparity test, segmented analysis and model de-bias through binning and thresholding techniques.
\end{enumerate}

\section{Future Plan}
PiML toolbox is our new initiative of integrating state-of-the-art methods in interpretable machine learning and model diagnostics. It provides convenient user interfaces and flexible APIs for easy use  of model interpretation, explanation, testing and comparison. Our future plan is to continuously improve the user experience, add new interpretable models, and expand the diagnostic suite. It is also our plan to enhance PiML experimentation with tracking and reporting. 

\section*{Acknowledgements}
We would like thank many people for their contributions and valuable comments to the development of PiML toolbox\footnote{See the name list in https://github.com/SelfExplainML/PiML-Toolbox/blob/main/Acknowledgements.md}. Our thanks also go to countless feedback from Model Risk Management communities of various financial institutions!
 
\bibliographystyle{apalike}
\bibliography{PiML}
	
\end{document}